\documentclass[letterpaper, 10 pt, conference]{ieeeconf}  
\usepackage{times}
\usepackage{epsfig}
\usepackage{graphicx}

\usepackage{amsthm}

\usepackage{mathrsfs}
\usepackage{graphics} 
\usepackage{epsfig} 
\usepackage{mathptmx} 
\usepackage{times} 
\usepackage{amsmath} 
\usepackage{amssymb}  
\usepackage{float} 
\usepackage{subfigure}
\usepackage{multirow}
\usepackage{enumerate}
\usepackage{amsfonts}
\usepackage{cite}

\IEEEoverridecommandlockouts                              

\title{\LARGE \bf
GPR-based Model Reconstruction System for Underground Utilities Using GPRNet
}

\author{Jinglun Feng$^{1}$, Liang Yang$^{1}$, Ejup Hoxha$^{1}$, Diar Sanakov$^{1}$, Stanislav Sotnikov$^{1}$, Jizhong Xiao*$^{1}$
\thanks{$^{1}$ Electrical Engineering Department, The City College of New York, New York, USA. 
        {\tt\small jfeng1,lyang1@ccny.cuny.edu}, 
        {\tt\small ehoxha000,ssanakov,sstonik@citymail.cuny.edu}, 
        the corresponding author is {\tt\small jxiao@ccny.cuny.edu}}%
}

\begin{document}

\maketitle
\thispagestyle{empty}
\pagestyle{empty}

\begin{abstract}
Ground Penetrating Radar (GPR) is one of the most important non-destructive evaluation (NDE) instruments to detect and locate underground objects (i.e. rebars, utility pipes). Many of the previous researches focus on GPR image-based feature detection only, and none can process sparse GPR measurements to successfully reconstruct a very fine and detailed 3D model of underground objects for better visualization. To address this problem, this paper presents a novel robotic system to collect GPR data, localize the underground utilities, and reconstruct the underground objects' dense point cloud model. This system is composed of three modules: 1) visual-inertial-based GPR data collection module which tags the GPR measurements with positioning information provided by an omnidirectional robot; 2) a deep neural network (DNN) migration module to interpret the raw GPR B-scan image into a cross-section of object model; 3) a DNN-based 3D reconstruction module, i.e., GPRNet, to generate underground utility model with the fine 3D point cloud. In this paper, both the quantitative and qualitative experiment results verify our method that can generate a dense and complete point cloud model of pipe-shaped utilities based on a sparse input, i.e., GPR raw data, with incompleteness and various noise. The experiment results on synthetic data as well as field test data further support the effectiveness of our approach.
\end{abstract}

\section{Introduction}
\begin{figure}[htbp]
\centering
    \subfigure[The omnidirectional robot for visual-based GPR data collection, where a GPR antenna is installed on the bottom of the robot chassis.]{
        \includegraphics[width=0.45\textwidth]{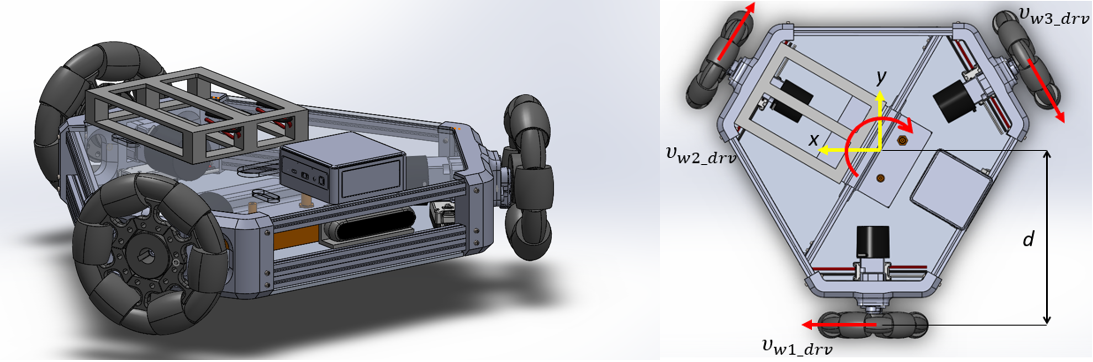}
    }
    \label{fig:intro_4}
    \quad
    \subfigure[Two different view angles of a concrete slab with two utility pipes buried in.]{
        \includegraphics[width=0.45\textwidth]{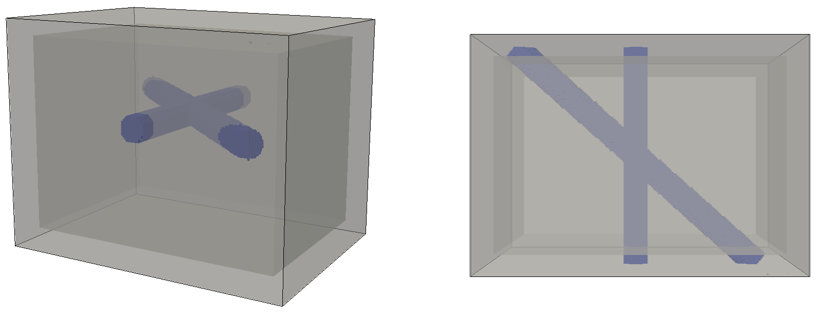}
    }
    \label{fig:intro_0}
    \quad
    
    \subfigure[The left part shows the GPR data collection results with B-scan images while the right part shows the GPR data interpretation results which indicate the cross-section of utility pipes.]{
        \includegraphics[width=0.45\textwidth]{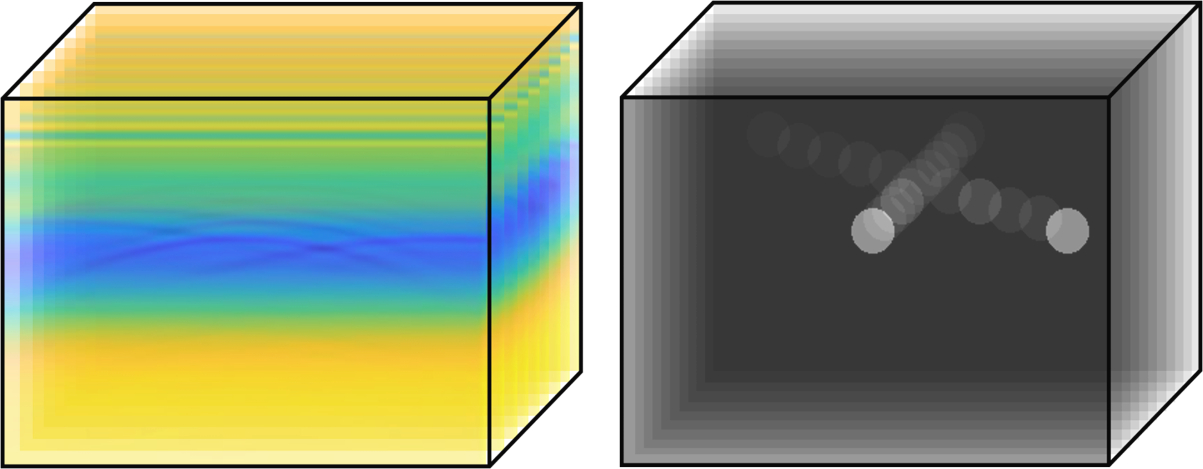}
    }
    \label{fig:intro_3}
    \quad
    
    \subfigure[The input indicates the coarse point cloud which represents the GPR interpretation results, while the output is a coarse-to-fine format by implementing our proposed GPR model reconstruction approach.]{
        \includegraphics[width=0.45\textwidth]{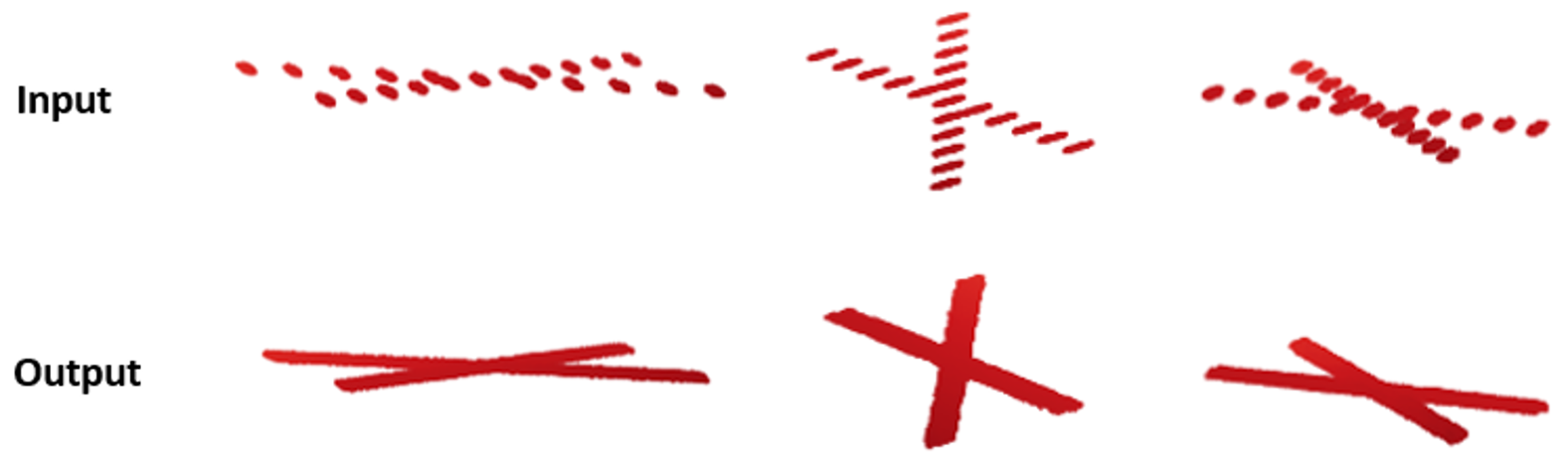}
    }
    \label{fig:intro_2}
    \quad

    \caption{Our proposed GPR-based point cloud model reconstruction system, where (a) demonstrates our robotics GPR data collection platform (b) shows the layout of underground utilities. GPR data is firstly collected with a loosely spacing distance. Then, by implementing a GPR interpretation method, we can convert GPR raw data into a cross-section image which matches the utilities, as depicts in (c). At last, by representing interpretation results into 3D point cloud format, (d) indicates that our proposed method could complete the 3D coarse point cloud input into a dense representation.}
    \label{fig:intro}
\end{figure}

Ground penetrating radar (GPR) has been widely used as a remote sensing technique in geophysical and civil engineering applications to provide effective radargram image and detection of underground structures. Among various types of none-destructive evaluation (NDE) techniques, such as acoustic, infrared, thermography \cite{demirci2012ground}, GPR gains its fame due to the high resolution capability and applicability in different fields detection. Specifically, GPR could not only evaluate the location and condition of underground utilities such as rebars, pipes and power cables, but also reveal the relative size of the subsurface objects.

Thanks to the advent of Deep Neural Networks applications in the GPR area, learning-based architectures are capable of directly operating on GPR data. DepthNet \cite{feng2020gpr} introduces the first deep-learning-based network for dielectric prediction of GPR data. Moreover, with the development in learning-based shape completion task, such as L-GAN \cite{achlioptas2018learning}, PCN \cite{yuan2018pcn}, FoldingNet \cite{yang2018foldingnet} and PF-Net \cite{huang2020pf}, it allows us to reconstruct 3D point cloud representation for underground objects.

However, in the practical GPR applications, the challenge of such a technique mainly lies in the following factors: data collection, GPR image interpretation, and GPR data visualization. Specifically, current GPR data collection requires inspector moves GPR device along grid lines without any rotations, which is hard to implement by regular ground mobile robots which cannot move sideway. As for the interpretation of the complex GPR data, it requires experienced professionals to interpret the GPR B-scan hyperbolic features with trained eyes. In addition, the reconstruction model produced by conventional migration algorithm contains too much noise and not easy to understand.

To address these challenges, we propose a novel 3D model reconstruction system based on GPR data collected by an omnidirectional robot to visualize the underground utilities, which is illustrated in Fig.\ref{fig:intro}. The key difference between our method and existing ones are that only a small amount of GPR data is required for 3D model reconstruction, in the meanwhile, it still achieves a better visualization result modelled by 3D point clouds. 

The main contributions of this work are:
\begin{itemize}
    \item an automated GPR data collection solution by using an omnidirectional robot, which holds the GPR antenna on the bottom and move forward, backward and sideways without any rotations to collect GPR data.
    \item a DNN model that operates and interprets the GPR B-scan image into a cross-section of underground objects model.
    \item a novel 3D reconstruction model that takes sparse GPR measurement as input and complete it with a dense point cloud which is able to present the fine details of the object structure.
\end{itemize}

\section{Related Works}

Our work is motivated by a number of related works which aim to interpret GPR radargram data and the coarse-to-dense 3D point clouds completion.  

\textbf{Mathematical Methods on GPR Data Interpretation:} The mathematical-based technique which serves as the traditional GPR imaging tools, enabling to convert the unfocused raw B-scan radargram data into a focused target, is called \emph{migration}. Migration methods can be roughly categorized into Kirchhoff migration, the phase-shift migration, the finite-difference method and back-projection algorithm. 

Kirchhoff migration method \cite{schneider1978integral} is first introduced in 1970s, \cite{liu2017ground} tests Kirchhoff migration method by using both synthetic and real GPR data, the results show that though the localization of the buried targets could be achieved, the information about target shape and extension are not provided by implementing this method.
In the meanwhile, \cite{smitha2016kirchhoff} represents that Kirchhoff migration method is capable to focus on the target position, however, it's processing speed is slower than the rest of migration approaches. 

Similar to Kirchhoff migration, phase-shift migration is first proposed and applied in 1978\cite{gazdag1978wave}, this is also a mathematical-based method which utilizes the ESM concept \cite{lecomte2005improving}. For comparable accuracy, this approach to migration is computationally more efficient than finite‐difference \cite{claerbout1972downward} approach, which is designed for the geometry of a single radargram source with a line of surface receivers.

Unlike the above conventional migration methods, the back-projection algorithm (BPA) back projects the emitted signal based on Electromagnetic waves travel path and the associated travel time, analogous to the traditional back-projection methods in computer aided tomography \cite{pereira20183d}. By implementing BPA, \cite{demirci2012study} conceals object detection through Millimeter-wave 2D imaging methods and works well both on metallic and dielectric types. BPA is also exploited for imaging of water leaks from buried pipes \cite{demirci2012ground}.

\textbf{Machine Learning Methods on GPR Data Interpretation:} In addition to the many researches that focus on the mathematical-based GPR migration methods, machine learning based methods are also widely studied to interpret GPR data, which could be catergorized into Hough transform method, SVM method and deep neural network (DNN) methods.

In 2000s, Hough transform method was first applied for underground targets detection \cite{al2000automatic}, however, this method is not effective on complex curves and diverse signals\cite{illingworth1988survey}. Furthermore, SVM applications play a important role in GPR studies on GPR imaging classification \cite{el2013material,ozkaya2020gpr} for underground materials.

Compared with conventional machine learning related methods, more and more DNN-based implementations  are applied in GPR data interpretation due to its higher accuracy and better efficiency capability, especially in detection, localization and classification of hyperbolic features in GPR radargram image \cite{liu2020detection,gao2020autonomous,khudoyarov2020three,feng2020gpr}.

\textbf{Deep Learning on 3D Point Cloud Completion:} Point cloud has been widely used in 3D vision related researches, which become one of the hottest topics in deep learning field. Specifically, PointNet and its following works \cite{qi2017pointnet,qi2017pointnet++,qi2018frustum} is the pioneer in this area and still be the state-of-the-art until now. It solves the variance of permutation and robustness exists in point clouds, by proposing a symmetric aggregation function and pointwise multilayer perceptrons. 

By taking advantage of the above works and \cite{yang2018foldingnet}, \cite{yuan2018pcn} proposes a point cloud based completion network, which combines the merits of PointNet and FoldingNet, to complete point cloud in a coarse-to-fine fashion. In addition, \cite{huang2020pf} presents a multi-stage completion loss to make a detailed completion in pointwise 3D shape. Moreover, a encoder-decoder architecture is designed in \cite{tchapmi2019topnet} to achieve the shape completion task. These methods all achieve a good result on both completion accuracy and noise robustness.

\section{Methodology}
In this section, we discuss the proposed GPR-based 3D model reconstruction system. Firstly, the vision-based robotic GPR data collection module provides a 6-DOF pose information of the GPR data, and it allows robot moves forward, backward and sideward without rotation during GPR data collection. Secondly, we interpret the GPR B-scan images into the cross-section of underground object model, and representing them as point cloud. Thirdly, a coarse 3D map is generated by registering the cross-section point cloud into the 3D space, where the 3D positioning information is obtained from the GPR data collection module. At last, we propose learning-based method to complete the sparse point cloud, thus provide dense model for better visualization of the underground utilities. 

\subsection{Visual-Inertial Robotic GPR Data Collection Module}
\label{section:vio}
To automate the GPR data collection, we developed an omnidirectional robot for the inspection of underground utilities. Our robot uses triangle-shaped chassis and three Mecanum wheels to move in any direction, it avoids rotation motion because it may cause the failure of vision-based positioning. As depicts in Fig.\ref{fig:intro}(a), our robot motion meets the following relation:

\begin{equation}
       \begin{bmatrix}
        v_{w1\_drv} \\
         v_{w2\_drv} \\
          v_{w3\_drv} \\
       \end{bmatrix} = 
       \begin{bmatrix}
       1 & 0 & -d \\
       \cos{\frac{2\pi}{3}} & \sin{\frac{2\pi}{3}} & -d \\
       \cos{\frac{-2\pi}{3}} & \sin{\frac{-2\pi}{3}} & -d \\
       \end{bmatrix}
       \begin{bmatrix}
        v_{x} \\
        v_{y} \\
        \omega \\
       \end{bmatrix}
        \label{eq:xNominal}
    \end{equation}
where $v_{w1\_drv}$, $v_{w2\_drv}$, $v_{w3\_drv}$ represents the linear velocity of each wheel, $d$ indicates the distance between the center of a wheel and the center of the robot body. $v_{x}$, $v_{y}$ and $\omega$ represent the linear velocity and angular velocity of the robot body respectively. 

In order to provide pose information for GPR B-scan data, the robot also carries an Intel NUC on-board computer and an Intel D435i RGB-D camera, which has a 6-axis IMU embedded in, to provide accurate and robust pose estimation for GPR sensor. We further fuse the IMU and the visual odometry data to improve the positioning accuracy by using a loosely-coupled approach, which was introduced in our previous work \cite{feng2020gpr}.

\subsection{GPR Data Analysis}

GPR antenna transmits a pulse of high-frequency electromagnetic (EM) wave in the ground \cite{kang2020deep} and waits for the EM waves echos. The signal emitted from GPR antenna is affected by EM characteristics, such as the dielectric of the subsurface objects or materials. 
\begin{figure}[H]
    \centering
    \includegraphics[width=0.35\textwidth]{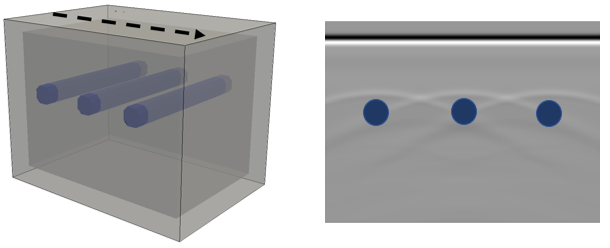}
    \caption{The CAD model in the left indicates a slab with three utilities buried in while the dot line on the top of the slab represents the motion direction of GPR device for data collection, which generates a B-scan image shows as in the right figure. The round shape features in the B-scan image indicates the cross-section of the utilities in the 3D model.}
    \label{fig:bscan-intro}
\end{figure}
As depicted in Fig.\ref{fig:bscan-intro}, the dotted line on the top of the slab which has three utilities embedded in, indicates the trajectory of a GPR device for data collection. Therefore, when GPR antenna scans along the perpendicular direction of the pipes, the composition of received signals can be identified as the hyperbolic feature because of the different dielectric between underground material and utilities. As shown in the right side of the Fig.\ref{fig:bscan_intro}, the image generated by a set of the transmit pulses is called \emph{B-scan}, where the around blue feature on the B-scan image indicates the cross-section of the utilities.

\subsection{GPR Data Interpretation}
\label{section:gpr_interpretation}

In this section, we introduce a learning model, which is a simple version of our previous work, MigrationNet\cite{Feng_2021_WACV}, to interpret subsurface objects. Compared with the existing learning-based method of GPR, our approach does not focus on the detection of hyperbola feature in the raw B-scan image, which is not helpful enough to GPR data interpretation task. MigrationNet aims at inferring the migration result by interpreting B-scan measurement.

\begin{figure}[H]
    \centering
    \includegraphics[width=0.48\textwidth]{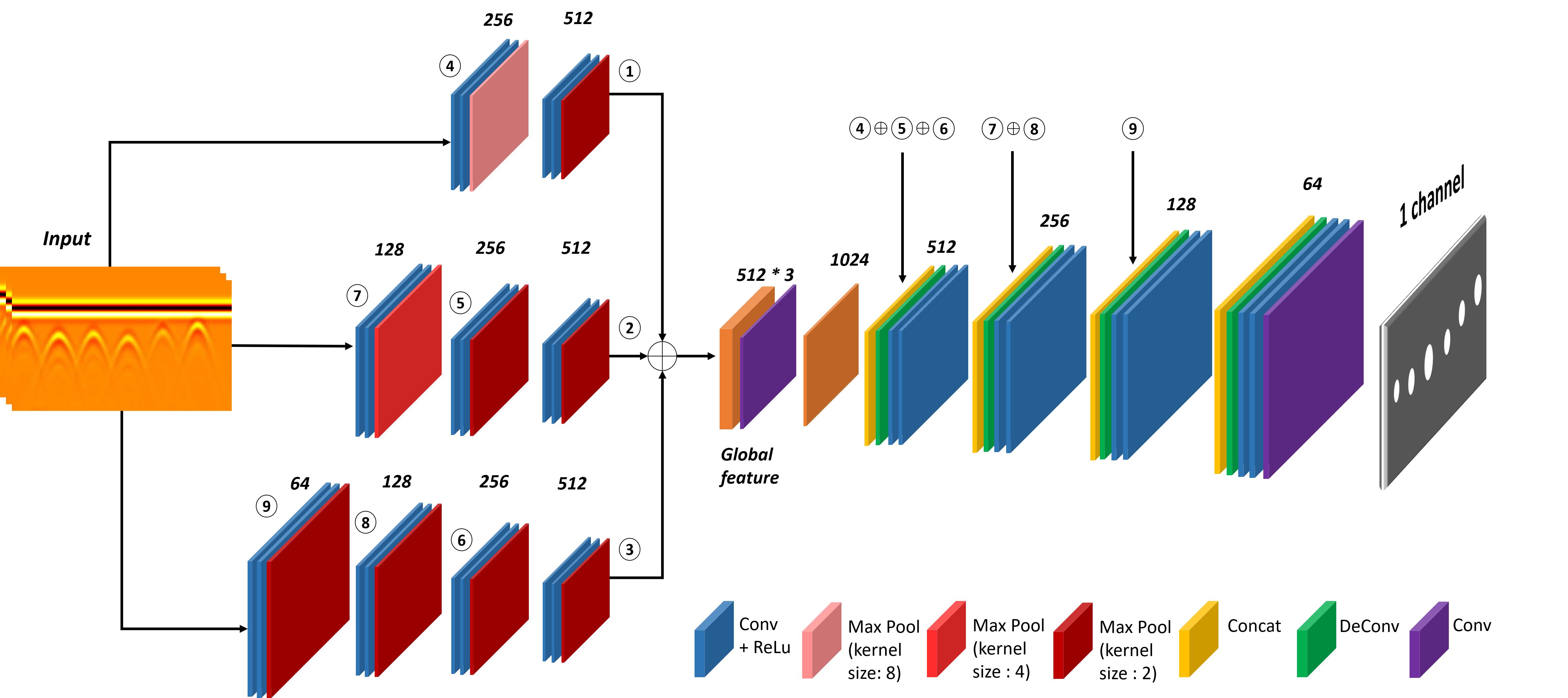}
    \caption{Framework of the MigrationNet. The input is the raw GPR B-scan image. Then, the features are extracted through the multiple resolution encoder and further concatenated into 1536 channels. The encoder consists of several de-convolutional groups, the global feature is produced by concatenating local features from MRE through skip-connection operation indicated by $\oplus$ while the numbers 1-9 indicate the layers which are used for concatenation, and finally decoded into a binary migration image.}
    \label{fig:bscan_intro}
\end{figure}

\begin{figure*}[ht]
    \centering
    \includegraphics[width=0.85\textwidth]{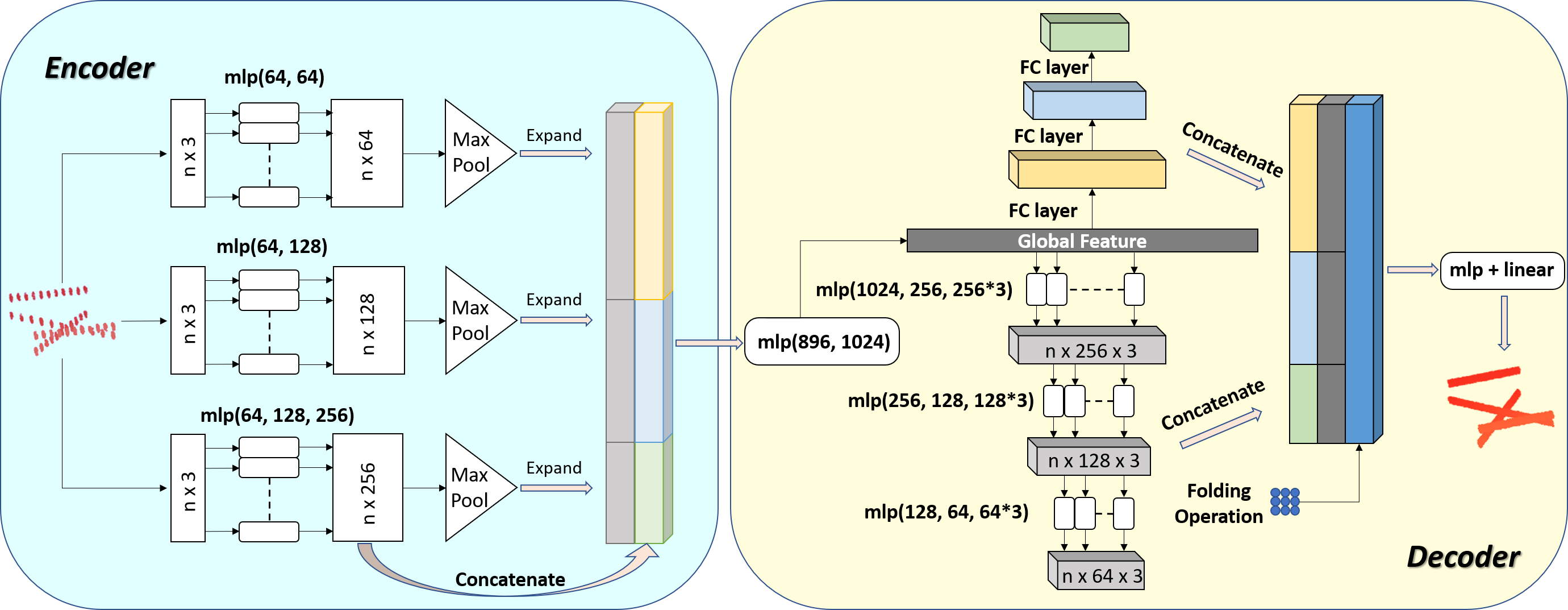}
    \caption{GPRNet framework. The input is a sparse point cloud which represents the cross-section of utilities, the encoder abstracts the input as a global feature and the decoder recovers the global feature to a dense point cloud.}
    \label{fig:schmatic}
\end{figure*}

It illustrated in Fig.\ref{fig:bscan_intro}, the input of MigrationNet is raw GPR B-scan data and it connects to our proposed Multiple Resolution Encoder (MRE).

The proposed encoder inherits the context capture ability by combining two convolution layers and one max-pooling layer. We follow the down-sampling group to encode the input data into a feature map with the same size, where size = [$M$ x $N$ x $512$], $M$ and $N$ indicates the height and width of GPR B-scan image respectively. In details, to get the same size of output feature, our input data is extracted into the different resolution by MRE. Specifically, at the top layer of MRE, the input data follows a down-sampling group where the kernel size of max-pooling layer is 8. In the middle layer, the kernel size of the first max-pooling layer is 4 while the rest of pooling layers' kernel size are all equal to 2. At the bottom layer, all the kernel size of max-pooling layers in the down-sampling groups are 2 and it allows the final output feature map has the same size in each input. At last, all three feature maps are then concatenated together as the global feature map, where size = [$M$ x $N$ x $1536$]. This design brings the combined latent feature to contain more detailed information of the input data.

The decoder takes global feature map as input to predict a [$M$ x $N$ x $1$] binary image, with the white indicates the pipe and the black indicates the background. In details, our decoder consists of 5 up-sampling group, and each group contains two convolutional layers and one deconvolutional layer. Besides, we also take the advantage of skip connections as illustrated in Fig.\ref{fig:bscan_intro}. 

\subsection{GPRNet for Underground Utilities Reconstruction}

Section.\ref{section:gpr_interpretation} facilitates to interpret raw B-scan into a cross-section image, however, MigrationNet only provides the slices of utilities' structure. Nevertheless, by taking advantages of the pose information obtained from Section.\ref{section:vio}, we could register the interpreted slices into the 3D space, to make up a sparse point cloud of the subsurface objects. Thus, we need to further complete the sparse point cloud map to recover the whole structure of utilities.

Inspired by the 3D point cloud completion works \cite{yang2018foldingnet,yuan2018pcn,tchapmi2019topnet,huang2020pf} in computer vision area, we proposed a model, i.e. GPRNet, to address the challenge mentioned above. We firstly represent the interpreted image, which indicates the cross-section of the underground utilities, as the 3D point cloud data format. Then, our encoder-decoder based network would complete the sparse input to generate a smooth and fine 3D point cloud reconstruction map.

\subsubsection{Encoder Design}

Our proposed encoder is an extended version of PCN \cite{yuan2018pcn}, it takes charge of representing the geometric information in the input point cloud as the global feature vector $\mathbf{v} \in \mathbb{R}^{n}$ where $n=896$. In addition, our encoder inherits the feature extraction ability by implementing PointNet layer, which is a combination of the convolutional multi-layer perceptron (MLP) layer and the max-pooling layer. Compared with the previous methods \cite{qi2017pointnet,qi2017pointnet++}, our encoder could extract multiple resolution information of the input data, which leads a better performance on small structure completion.    

Specifically, by passing through three MLP layers which have the different dimensions, our input $m \times 3$ point cloud data, where $m$ is the number of the points and set to $1500$, is firstly encoded into three point feature vectors $\mathbf{f}_{i}$, where size $\mathbf{f}_{i} := $ $m\times64$, $m\times128$, $m\times256$ for $ i = 1, 2, 3$ respectively. Then, a max-pooling layer is performed on each extracted feature to obtain three intermediate features $\mathbf{g}_{j=1,2,3}$ with multiple dimensions $[64 - 128 - 256]$. Furthermore, we firstly concatenate each point feature vector $\mathbf{f}_{i}$ and each intermediate feature $\mathbf{g}_{j}$ together to obtain a expanded feature, which includes the feature information at different level. In addition, each $\mathbf{g}_{j}$ is concatenated to each $\mathbf{f}_{i}$ to obtain the feature matrix $\mathbf{F}$. At last, $\mathbf{F}$ is passed through another PointNet layer to generate the global feature vector $\mathbf{v}$.

\subsubsection{Decoder Design}

Fully-connected decoder \cite{achlioptas2018learning} is good at predicting the global geometry of point cloud, however, it ignores the local features. FoldingNet decoder \cite{yang2018foldingnet} is good at generating a smooth local feature. Therefore, by combining the above decoders, we design our decoder as a hierarchical structure \cite{huang2020pf}. 

In the primary decoder layer, the global feature vector is extracted into three feature vectors (size:=256,128,64 neurons respectively) by passing through the fully-connected layers, which is responsible for generating point cloud in different resolutions. Then, these three vectors are expanded into three local feature matrices (size:= $256 \times 3$,$128 \times 3$,$64 \times 3$). In the meanwhile, in order to get a better structure sense from global feature, the global vector is reshaped into the same size of the local feature matrices through the MLP layers, as it is shown in  Fig.\ref{fig:schmatic}. Then, the global feature and local feature are concatenated together as a $896 \times 3$ matrix. At last, by taking advantages of folding operation, a patch of $9$ points is generated at each point in the point set generated from last step. Thus, we can obtain the detailed output consisting of $896*9$ points. A dense point cloud output is then generated from our multi-resolution decoder, based on the fully-connected and folding operations.

Thanks to this multi-resolution architecture, high-level features will affect the expression of low-level features, and low resolution features can contribute to form the global feature, which provide a sense of local geometry of the shape. Our experiments show that the prediction of our proposed decoder has fewer distortions and better detailed geometry of the shape.

\subsubsection{Loss Design}

To constrain and compare the difference between the output point cloud $\textit{S}$ and the ground truth point cloud $\textit{S}_{gt}$, an ideal loss must be differentiable with respect to point locations and invariant to the permutation of point cloud. In this paper, we use Chamfer Distance (CD), which is proposed by Fan \emph{et al.} \cite{fan2017point}. It calculates the average closest point distance between $\textit{S}$ and $\textit{S}_{gt}$, that meets the requirement of the above conditions:

\begin{equation}
d_{CD}(\textit{S},\textit{S}_{gt}) = \frac{1}{\textit{S}} \sum_{x \in \textit{S}} \min \limits_{y \in \textit{S}_{gt}}\|x-y\|_{2} + \frac{1}{\textit{S}_{gt}} \sum_{y \in \textit{S}_{gt}} \min \limits_{x \in \textit{S}}\|y-x\|_{2}
\label{equ:loss}
\end{equation}

The Chamfer Distance finds the nearest neighbor in the ground truth point set, thus it can force output point clouds to lie close to the ground truth and be piecewise smooth.

\section{Experimental Study}
To evaluate our approach, we perform $725$ tests on NDT-GPR dataset \cite{feng2020gpr} while $120$ automated field tests using GPR-Cart. The field tests of robotic GPR inspection system was conducted on a concrete slab at River Edge, NJ, USA. In this section, the effectiveness and robustness of our 3D reconstruction method as well as the results of on-site tests are discussed in details. 

\subsection{Model Training and Evaluation}

\textbf{Dataset:}
In this paper, we use synthetic GPR dataset proposed in \cite{feng2020gpr}. This dataset builds a synthetic testing environment which simulates the real NDT condition, where rebars, utilities and PVC pipes are buried underneath the surface. This simulated environment mimics this property and involves pipe-shaped objects with different location as well as the size. Noticed all of the simulated objects have a round cross section. Specifically, this dataset contains $1628$ B-scan data which could convert into images, as well as their cross section images as the ground truth. After register each cross-section image into the 3D point cloud, each point set contains $1500$ points while each ground truth point set contains $8064$ points.

\textbf{Network Training:}
We train our model on a server with Intel Core i9-9900K 3.2GHz CPU, GeForce RTX 2080 Ti GPU, and 32GB RAM. Among all the $1628$ model of GPR data, we reserve $100$ models for validation and $150$ models for testing, the rest is used for training. Notice that our models are trained for $100$ epochs with an Adam optimizer. The initial learning rate is set to 0.00005 while the batch size is $16$. The weight decay is $0.7$ for every 50000 iterations. 

\subsection{Model Reconstruction Study}

\textbf{Baseline Comparison:}
In order to assess the effectiveness of our proposed method, we compare our baseline against PCN and PF-Net. Since related point cloud completion works are trained with different datasets, we use the GPR B-scan dataset \cite{feng2020gpr} which matches our testing condition so that we can evaluate the above methods quantificationally. Specifically, we use two evaluation metrics to compare the methods mentioned above, which are the average squared distance \cite{achlioptas2018learning} and the $L_{1}$ distance from each point between output point cloud and the ground truth. It need to be noted that the number of points in the predicted output is set to $8192$ while the input number of points are $1500$.

\begin{figure*}
    \centering
    \includegraphics[width=0.9\textwidth]{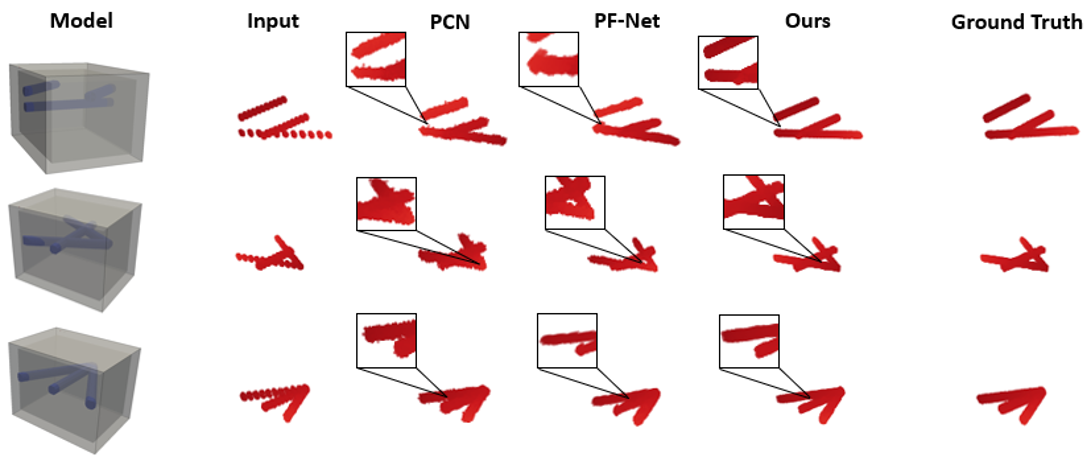}
    \caption{The comparison of completion results between other methods and our network. From left to right: the slab CAD model, input data, PCN\cite{yuan2018pcn}, PF-Net\cite{huang2020pf}, our method and the ground truth. Based on the results, our method could reconstruct a better 3D model for visualization.}
    \label{fig:pcn_result}
\end{figure*}

\begin{table}[!th]
\caption{Evaluation Performance Comparison with Different Baselines. $CD$: average squared distance between two points, $L_{1}$: norm distance between two points}
\label{table:weight}
\begin{center}
\begin{tabular}{|c|c|c|c|c|c|c|c|c|}
\hline
\hline
&\multicolumn{2}{|c|}{GPRNet} &\multicolumn{2}{|c|}{PCN} &\multicolumn{2}{|c|}{PF-Net}\\ 
\hline
\hline
CD & \multicolumn{2}{|c|}{6.214}&\multicolumn{2}{|c|}{6.965} &  \multicolumn{2}{|c|}{{7.005}} \\
\emph{L}1 & \multicolumn{2}{|c|}{22.97} &\multicolumn{2}{|c|}{24.90} &  \multicolumn{2}{|c|}{{33.68}} \\
\hline
\end{tabular}
\end{center}
\label{Table:baseline}
\end{table}

The results in Table \ref{Table:baseline} shows that our proposed method outperforms other methods in different GPR-based inspection environments, note that the Chamfer distance is reported multiplied by $10^3$ while $L_{1}$ distance is scaled by $100$. Note that both the errors are evaluated by the united distance. In addition, we could also visualize the output utilities point cloud generated by all the methods. Compared to other methods, the prediction of our method outperforms the other methods in spatial continuity and shape accuracy level. 
\begin{figure}[H]
    \centering
    \includegraphics[width=0.47\textwidth]{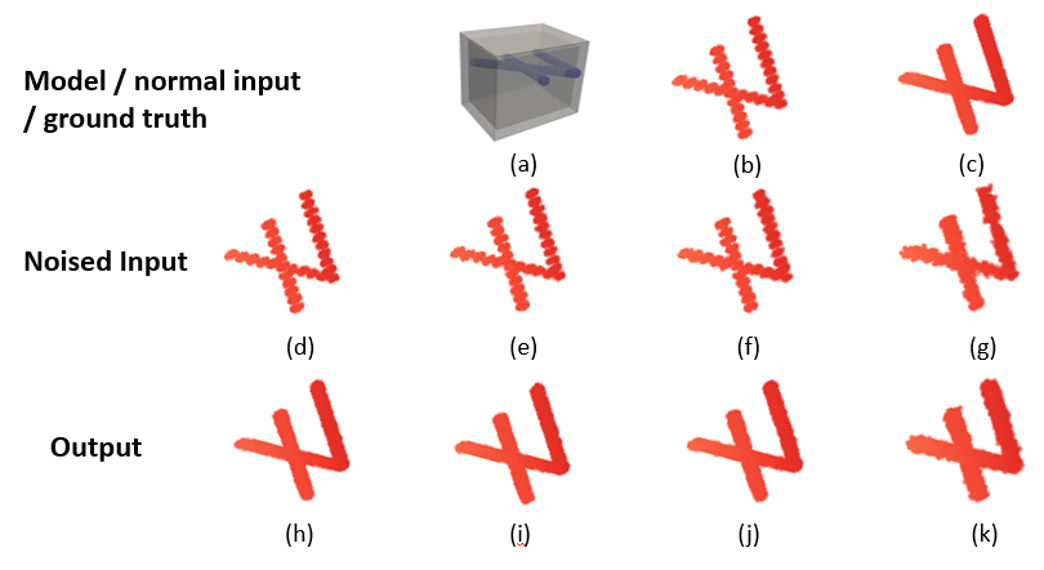}
    \caption{Noise robustness comparison result. Pictures (a)-(c) indicate the model, raw input and ground truth respectively. The second line represents the Gaussian-white-noised input, where noise variance are $0.01$, $0.05$, $0.1$, $0.2$ respectively, while pictures (h)-(k) in the last line demonstrate the predicted results of the noised input (d)-(g).}
    \label{fig:noise_comparision}
\end{figure}

\begin{table}[!th]
\caption{Noise Robustness Evaluation on GPRNet with two metrics.}
\label{table:NosieComparision}
\begin{center}
\begin{tabular}{|c|c|c|c|c|c|c|}
\hline
\hline
& \multicolumn{2}{|c|}{CD distance} & \multicolumn{2}{|c|}{\emph{L1} distance} \\ 
\hline
\hline
Variance \& Noise density = 0.01  &  \multicolumn{2}{|c|}{6.350}  & \multicolumn{2}{|c|}{23.029}   \\
Variance \& Noise density = 0.05 & \multicolumn{2}{|c|}{6.715} & \multicolumn{2}{|c|}{24.851}   \\
Variance \& Noise density = 0.1 &  \multicolumn{2}{|c|}{7.374} &  \multicolumn{2}{|c|}{25.639}  \\
Variance \& Noise density = 0.2 & \multicolumn{2}{|c|}{7.765} & \multicolumn{2}{|c|}{26.589}   \\
\hline
\end{tabular}
\end{center}
\end{table}

\begin{figure*}[htbp]
\centering
    \subfigure[The ground truth of the concrete slab.]{
        \includegraphics[width=0.31\textwidth]{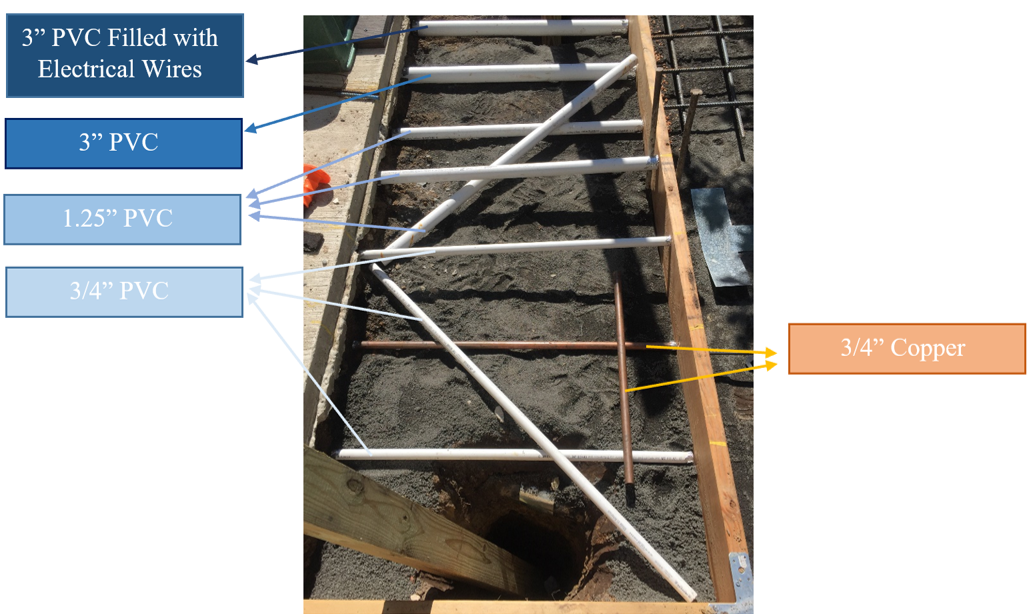}
    }
    \label{fig:field_slab}
    \subfigure[The design details of the concrete slab.]{
        \includegraphics[width=0.31\textwidth]{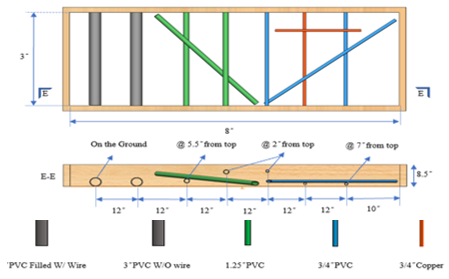}
    }
    \label{fig:field_slab}
    \subfigure[3D reconstruction result of the field test.]{
    	\includegraphics[width=0.31\textwidth]{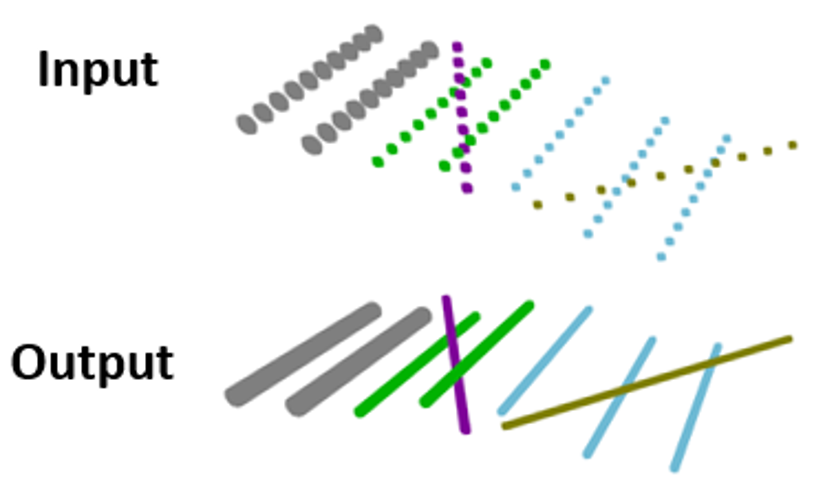}
    
    }
    \label{fig:field_test}
    \quad
    \caption{A field test on a concrete slab which has multiple utilities embedded in, as demonstrated in (a) and (b). Our robot collected 10 groups of GPR data along the straight lines perpendicular to the pipes. As shown in (c), our input data is coarse and represents the cross-section of the utilities while the predicted result indicates the 3D model of the embedded pipes.}
    \label{fig:filed_test}
\end{figure*} 

\textbf{Noise Robustness:}
To evaluate the effectiveness of our method under different sensor noise level, we perturbed the input sparse point cloud with multiple Gaussian white noise levels, where the standard deviations are 0.01, 0.05, 0.1 and 0.2 respectively. For each data set, we further compare average squared distance and $L_{1}$ distance between the noised data input and ground truth. As illustrated in Table.\ref{table:NosieComparision} and Fig.\ref{fig:noise_comparision}, we could find our proposed method achieved higher robustness against the noise.

\subsection{Dataset Test and Field Test}
In this section, we evaluate the robotic GPR inspection system with both synthetic dataset and field test. The performance of underground utilities reconstruction is illustrated in Fig.\ref{fig:pcn_result}, where the model of the synthetic slabs, coarse input point cloud data, predicted output data and the ground truth are provided. The field test result is also demonstrated in Fig.\ref{fig:filed_test}, where pipes with different size are colored respectively. In this test, we drive our omnidirectional robot with a sparse line spacing, and record the pose information with our visual-inertial module. Specifically, our robot moves forward, backward and sideways without any rotations, to collect GPR data along its trajectory. By comparing the reconstructed 3D map with the ground truth of the slab in our test pit, the effectiveness of proposed system is well verified.

\section{Conclusion}
This paper presents a novel GPR-based 3D model reconstruction system for underground utilities by using an omnidirectional robot. Our omnidirectional robot allows GPR device move forward, backward and sideward without rotations. The proposed system is able to reconstruct underground utilities model represented as 3D point cloud. According to the experiments, our approach could obtain a fine 3D model for a better visualization while we could find out our method has a higher robustness against noise. 

\section{Acknowledgement}

Financial support for this study was provided by NSF grant IIP-1915721, and by the U.S. Department of Transportation, Office of the Assistant Secretary for Research and Technology (USDOT/OST-R) under Grant No. 69A3551747126 through INSPIRE University Transportation Center (http://inspire-utc.mst.edu) at Missouri University of Science and Technology. The views, opinions, findings and conclusions reflected in this publication are solely those of the authors and do not represent the official policy or position of the USDOT/OST-R, or any State or other entity. J Xiao has significant financial interest in InnovBot LLC, a company involved in R\&D and commercialization of the technology.  

\bibliographystyle{IEEEtran}
\bibliography{ref}

\end{document}